\documentclass[11pt,a4paper]{article}

\usepackage[margin=1in]{geometry}
\usepackage{microtype}
\usepackage{setspace}
\usepackage[T1]{fontenc}
\usepackage{lmodern}

\usepackage{amsmath,amssymb,amsfonts}

\usepackage{graphicx}
\graphicspath{{figures/}}
\usepackage{booktabs}
\usepackage{multirow}
\usepackage{array}
\usepackage{tabularx}
\usepackage{caption}
\usepackage{subcaption}

\usepackage[colorlinks=true,linkcolor=NUpurple,citecolor=NUpurple,urlcolor=NUpurple]{hyperref}
\usepackage{url}
\usepackage{xurl}
\usepackage[capitalize,noabbrev]{cleveref}
\usepackage[numbers,sort&compress]{natbib}

\usepackage{xcolor}
\definecolor{NUpurple}{HTML}{4E2A84}

\usepackage{listings}
\title{\textbf{HumanCentric-Scenes-AI}\\[6pt] \large A Human-Curated Multimodal Dataset of AI-Generated Human-Centric Scenes Across Surveillance, Occupational Portraits, AI-Edited Images, and Short-Form Video}

\author{
  Nima Kamali\thanks{University of Applied Science and Technology (Allameh Tabataba'i Center), Iran. \texttt{nima.kamaliz97@gmail.com}} \\
  \and
  Sana Nasiri\thanks{University of Applied Science and Technology (Allameh Tabataba'i Center), Iran. \texttt{nasirisana00@gmail.com}} \\
  \and
  Negar Kamali\thanks{Northwestern University. \texttt{negar.kamali@u.northwestern.edu}. Corresponding author.}
}

\date{\today}

\begin{document}
\maketitle

\begin{abstract}
\noindent
We introduce HumanCentric-Scenes-AI, a human-curated dataset of 296
AI-generated stimuli that depict people across four contexts in which
people encounter AI-generated media. The Surveillance subset contains 85
AI-generated CCTV-style images. The Occupations subset contains 128
AI-generated occupational portraits in a balanced paired design covering
64 occupations with one female and one male version of each. The
AI-edited subset contains 27 real photographs in which a region was
altered or replaced by a generative model. The Video subset contains 56
AI-generated short videos. We curated the images and videos manually over
multiple rounds to make them as photorealistic as possible. For each
stimulus we release the file, the generation prompt where recorded, and
per-stimulus metadata (\Cref{tab:metadata-schema}). We report depicted-gender distributions
from an initial set of 134 generations without a specified gender and compare them with U.S.\ Bureau of
Labor Statistics occupational gender shares. We report edit-region size
and its association with zero-shot detection difficulty for the AI-edited
subset. We report image-quality distributions for the Surveillance subset
and subject-category coverage for the Video subset.
HumanCentric-Scenes-AI is intended first for human-perception research,
and also for calibration of AI-detection classifiers and bias analysis of
generative models.
\medskip
\noindent The dataset is released under a CC BY-NC 4.0 license covering
the AI-generated stimuli and the metadata.
\begin{center}
\small\url{https://huggingface.co/datasets/Nima0Kamali/humancentric-scenes-ai}
\end{center}
\end{abstract}

\section{Introduction}\label{sec:intro}

Generative image and video models now produce scenes of people that
viewers often cannot distinguish from photographs
\citep{nightingale2022faces, lu2023seeing, kamali2025characterizing}.
Detection research has responded with large benchmarks of AI-generated
media. Widely used examples consist of face swaps and reenactments in
otherwise authentic video \citep{rossler2019faceforensics, li2020celebdf,
dolhansky2020dfdc} or of whole-image generations from many generators
\citep{zhu2023genimage, park2025community}.

We focus on four contexts in which people encounter AI-generated depictions of people. Surveillance footage is low-resolution and heavily
compressed, and the forensic traces that diffusion-image detectors rely
on weaken under compression and resizing \citep{corvi2023detection}.
Occupational portraits carry gender associations that generative models
are known to amplify \citep{bianchi2023easily, guilbeault2024online}.
Partially manipulated photographs alter only a region of an authentic
image and leave most pixels untouched \citep{mareen2024tgif}. Short-form video is a common format on social media, and several commercial generators now produce it.

We release HumanCentric-Scenes-AI, a dataset of 296 AI-generated stimuli
that covers these four domains (\Cref{tab:composition}). The Surveillance
subset contains 85 AI-generated CCTV-style images. The Occupations subset
contains 128 AI-generated occupational portraits in a balanced paired
design covering 64 occupations with one female and one male version of
each. The AI-edited subset
contains 27 AI-edited derivatives of public-source photographs. The Video
subset contains 56 videos from Veo, Sora and Kling. Every stimulus
carries its generation prompt where recorded and structured metadata
(\Cref{tab:metadata-schema}). We curated every stimulus manually, selecting for photorealism.
Earlier work found that human curation raises the photorealism of
AI-generated images \citep{kamali2025characterizing}. The dataset is
built for research on human perception of AI-generated media.

We find that Midjourney images prompted toward CCTV degradation remain sharper than
real CCTV frames (\Cref{sec:cctv}). We find that Midjourney depicts a man
in 73.9\% of generations under gender-neutral occupational prompts and
that descriptive phrasing softens the skew (\Cref{sec:occupation}). In the 25 characterized AI-edited images, larger edits are
associated with higher zero-shot classification accuracy
(\Cref{sec:partial}). The Video subset spans six subject categories, and
the characterized videos have median durations similar to those of their
real references (\Cref{sec:video}).

\paragraph{Contributions.} We release a four-domain multimodal
dataset of 296 AI-generated stimuli in full, with per-stimulus metadata
for every file (\Cref{tab:metadata-schema}). We characterize each domain
(\Cref{sec:cctv,sec:occupation,sec:partial,sec:video}) against third-party
statistical references where applicable, such as U.S.\ BLS occupational
gender shares for the Occupations subset. We additionally analyze an initial set of 134 generations without a specified gender across 67 occupations, which shows the generator default that the balanced release set cannot show.

\section{Related Work}\label{sec:related}

Several widely used deepfake detection datasets consist of faces
manipulated in otherwise unaltered video. FaceForensics++ \citep{rossler2019faceforensics},
Celeb-DF \citep{li2020celebdf} and the DeepFake Detection Challenge
dataset \citep{dolhansky2020dfdc} are widely used benchmarks for face-swap and reenactment forensics. Deepfake-Eval-2024 extends this line to manipulated media flagged in the wild across platforms and languages \citep{chandra2025deepfakeeval}. Earlier work by Korshunov and Marcel
\citep{korshunov2018deepfakes} framed face swaps as a threat to biometric
verification. In FaceForensics++, Celeb-DF and DFDC, a face is replaced or
reanimated while the rest of the scene is held fixed.

Other datasets contain fully AI-generated images. GenImage
\citep{zhu2023genimage} pairs over a million generated images with real
counterparts across eight generators, targeting classifier generalization
across architectures. Community Forensics scales generator coverage to several
thousand models to improve detector generalization
\citep{park2025community}. Corvi et al.\ show that the forensic traces
such detectors learn weaken under the compression and resizing that
shared media undergo \citep{corvi2023detection}. Our Surveillance subset
contains AI-generated images prompted toward such degraded capture
conditions, and we find that they remain sharper than real CCTV frames
(\Cref{sec:cctv}). For partial manipulation, TGIF releases roughly
80{,}000 text-guided inpainting forgeries for detector training
\citep{mareen2024tgif}. Our AI-edited subset is a small curated set built
for human-perception research, and in it larger edits are associated with
higher zero-shot classification accuracy (\Cref{sec:partial}).

Several studies measure demographic bias in text-to-image models. Bianchi
et al.\ \citep{bianchi2023easily} show that simple occupational prompts
amplify demographic stereotypes at scale, and DALL-Eval
\citep{cho2023dalleval} probes social biases across generation models.
Stable Bias quantifies demographic representation in diffusion output
\citep{luccioni2023stablebias}. Paired stereotype testing reports
gender-occupational bias in over 74\% of DALL-E~3 generations
\citep{wan2025maleceo}. Our Occupations subset contributes a complementary design. The same 67
occupations are prompted twice, once by name and once by paraphrased
description. This design separates the generator's male default (82.1\% under
occupation-name prompts) from a prompt-phrasing effect of 16.4 percentage
points (\Cref{sec:occupation}).

Our dataset differs from these datasets in purpose. It was curated as stimulus material for human-perception research,
extending the curation practice of our earlier work
\citep{kamali2025characterizing} to four domains. The release is small, and each stimulus carries structured metadata. It follows the datasheet
reporting framework of Gebru et al.\ \citep{gebru2021datasheets}
(\Cref{app:datasheet}).

\section{Dataset Composition}\label{sec:composition}

\begin{table}[t]
  \centering
  \small
  \caption{HumanCentric-Scenes-AI releases all 296 AI-generated
  stimuli.}
  \label{tab:composition}
  \begin{tabular}{l l l r}
    \toprule
    \textbf{Subset} & \textbf{Domain} & \textbf{Generator(s)} & \textbf{Released} \\
    \midrule
    Surveillance & CCTV-style images          & Midjourney                          & 85  \\
    Occupations  & Gender--occupation pairings & Midjourney                          & 128 \\
    AI-edited    & Partial manipulation       & Midjourney inpainting               & 27  \\
    Video        & Short-form video           & Veo 35, Sora 18, Kling 1, unattr.\ 2 & 56 \\
    \midrule
    \textbf{Total} & & & \textbf{296} \\
    \bottomrule
  \end{tabular}
\end{table}

The dataset is structured as four named subsets of AI-generated stimuli.
\Cref{tab:composition} gives the headline counts for the full release.

\subsection{Curation procedure}
Curators reviewed every candidate generation manually over multiple
rounds and kept those without visible artifacts that would make them
immediately distinguishable from photographs.
All images in the Surveillance and Occupations subsets were generated
with Midjourney. We generated the CCTV-style set with degradation-oriented
prompts (\Cref{sec:cctv}). We generated the gender--occupation set as a balanced paired design in
which each occupation is prompted once per gender (\Cref{sec:occupation}). The AI-edited subset applies AI edits to regions of
public-source photographs, released as derivative works. Video-subset stimuli were generated with Veo, Sora and Kling from
category-matched prompts. Each AI video was developed against a real
reference clip from the same subject category.

\subsection{Metadata schema}
\Cref{tab:metadata-schema} lists the columns of \texttt{metadata.csv}.

\begin{table}[t]
  \centering
  \small
  \caption{Schema of \texttt{metadata.csv} (one row per released file).}
  \label{tab:metadata-schema}
  \begin{tabular}{l l}
    \toprule
    \textbf{Column} & \textbf{Description} \\
    \midrule
    file\_name          & Repository path to the media file \\
    filename            & Content-hash filename (stable across releases) \\
    original\_filename  & Descriptive curation-time name (e.g., \texttt{go\_f\_054}) \\
    modality            & image / video \\
    label               & ai\_generated \\
    generator           & midjourney / veo / sora / kling / unknown \\
    generator\_version  & Generator version where recorded \\
    prompt              & Generation prompt where recorded \\
    category            & Domain tag (e.g., cctv, occupational\_portrait) \\
    subcategory         & Domain-specific tag (e.g., surveillance, gender\_balanced) \\
    occupation          & Occupations subset only \\
    depicted\_gender    & Occupations subset only (M / F) \\
    pose                & Where recorded \\
    source\_url         & Source link, where recorded \\
    contributor         & Curator responsible for the stimulus \\
    stimulus\_set       & Subset identifier (e.g., \texttt{set2\_cctv}) \\
    \bottomrule
  \end{tabular}
\end{table}

\section{Low-Resolution and CCTV Images}\label{sec:cctv}

The Surveillance subset comprises all 85 Midjourney CCTV-style images
generated for the project. The images were prompted to mimic CCTV captures
with their noise, blur and downsampling. The prompts specify Midjourney version 7
for 23 images, version 6.1 for 6 and version 6 for 3, and the remaining
53 prompts do not specify a version. The characterization below compares a
25-stimulus subset with 26 real CCTV frames from the CCTV Human Pose
Estimation Dataset~\citep{KaggleCCTVData}.

\subsection{Composition and metadata}
Image-quality statistics below are computed on the 25-image experimental
subset and its paired real references; extending them to the full 85-stimulus
release is a one-line change to the released analysis script.

\begin{table}[t]
  \centering
  \small
  \caption{Surveillance-subset image characteristics for the 25-stimulus characterization subset and its paired real references (the full release contains 85 Midjourney images). All stimuli were normalized to
  $600 \times 600$\,px square format during curation. Despite degradation-oriented
  prompts (``grainy low-resolution CCTV footage, 1990s video compression''),
  Midjourney generations are sharper than real CCTV frames on this
  measure. The median Laplacian variance is 649 for Midjourney and 212 for
  real frames (Mann--Whitney $U = 482$, $z = 3.29$, $p = 0.001$).}
  \label{tab:set2-resolution}
  \begin{tabular}{l r r r}
    \toprule
    \textbf{Measure (median [range])} & \textbf{Midjourney AI ($n{=}25$)} & \textbf{Real CCTV ($n{=}25$)} \\
    \midrule
    Resolution (px)            & $600 \times 600$      & $600 \times 600$ \\
    File size (KB)             & 128 [77--203]         & 98 [67--157] \\
    Sharpness (Laplacian var.) & 649 [31--9898]        & 212 [25--2383] \\
    \bottomrule
  \end{tabular}
\end{table}

\subsection{Generation prompts}

\begin{lstlisting}[caption={Representative Midjourney prompt used for Surveillance-subset generations.}]
grainy low-resolution CCTV footage, 1990s video compression,
digital noise, inside of a parking security camera, VHS, blurry,
low fps --v 6
\end{lstlisting}

\section{Images with Gender--Occupation Pairings}\label{sec:occupation}

The Occupations subset comprises 128 Midjourney occupational portraits in
a balanced paired design. Each of 64 occupations is prompted once with a
female subject and once with a male subject (e.g., ``portrait of a female
model on catwalk, shot on Kodak, grainy''). The balance is by construction, so the released subset
supports studies of occupation--gender congruence. The generator's
unconstrained gender prior is observable in the initial set of generations without a specified gender
instead.

\subsection{Gender distribution under gender-neutral prompts}

Before building the balanced release set, the same 67 occupations were
generated once per prompt style with no gender specified, yielding a
initial set of 134 generations without a specified gender. This initial set is analyzed here and is not part of the release. Under
these gender-neutral prompts, Midjourney depicted a man in 73.9\% of
generations (\Cref{tab:set3-contingency}). The share is 82.1\% when the prompt names the occupation and 65.7\% when it
describes the role. The direction is consistent with bias audits of other
text-to-image models \citep{bianchi2023easily, luccioni2023stablebias,
wan2025maleceo}. Eleven of the 67 occupations are rendered female under occupation-name
prompts. Seven of the eight with a published BLS share are female-majority
occupations in U.S.\ labor statistics (\Cref{tab:set3-stereotype-index}).

\begin{table}[t]
  \centering
  \small
  \caption{Depicted gender in Midjourney generations under gender-neutral
  prompts (initial set without a specified gender; $N = 134$ generations across 67
  occupations, one generation per occupation and prompt style). Two-sided exact binomial
  tests against $P(\text{male}) = 0.5$ exclude gender-unclear generations.}
  \label{tab:set3-contingency}
  \begin{tabular}{l r r r r l}
    \toprule
    \textbf{Prompt style} & \textbf{$N$} & \textbf{Male} & \textbf{Female} & \textbf{Unclear} & \textbf{Binomial $p$} \\
    \midrule
    Explicit occupation name     & 67  & 55 (82.1\%) & 11 (16.4\%) & 1 & $3.6 \times 10^{-8}$ \\
    Descriptive (role described) & 67  & 44 (65.7\%) & 22 (32.8\%) & 1 & $9.2 \times 10^{-3}$ \\
    \midrule
    \textbf{Overall}             & 134 & 99 (73.9\%) & 33 (24.6\%) & 2 & \\
    \bottomrule
  \end{tabular}
\end{table}

\subsection{Gender-stereotype congruence}

Occupations carry measurable gender associations. Guilbeault et
al.~\citep{guilbeault2024online} quantify them as a gender association
score derived from the proportion of male and female faces in online
images for an occupation, where $-1$ indicates fully female
representation and $+1$ fully male. The released metadata supports the
same congruence analyses, since every Occupations stimulus carries its
occupation and depicted gender. We treat a generation as
stereotype-congruent when its depicted gender matches the
female-majority or male-majority direction of the occupation in U.S.\
labor statistics. Of the eleven occupations that Midjourney renders female under
occupation-name prompts, seven are female-majority in BLS data,
newscaster is male-majority, and model, florist and cheerleader have no
published BLS share (\Cref{tab:set3-stereotype-index}).

\begin{table}[t]
  \centering
  \small
  \footnotesize
  \caption{Occupations rendered female under gender-neutral occupation-name
  prompts are predominantly female-majority occupations in U.S.\ labor
  statistics. The table lists all 11 of 67 occupations for which Midjourney
  depicted a woman under occupation-name prompting, alongside the percent
  of employed women in the closest matching detailed occupation from the
  U.S.\ Bureau of Labor Statistics Current Population Survey (2025 annual
  averages, Table cpsaat11)~\citep{bls2025cpsaat11}. Across all occupations,
  women are 47.1\% of the employed. Seven of the eight occupations with a
  published share exceed this baseline. Newscaster is the exception at
  41.1\%.}
  \label{tab:set3-stereotype-index}
  \begin{tabular}{l l r}
    \toprule
    \textbf{Occupation (MJ renders female)} & \textbf{BLS detailed occupation} & \textbf{\% women} \\
    \midrule
    Nurse              & Registered nurses                                    & 87.3 \\
    Dental hygienist   & Dental hygienists                                    & 95.0 \\
    Hairdresser        & Hairdressers, hairstylists, and cosmetologists       & 92.0 \\
    Assistant          & Secretaries and admin.\ assistants\textsuperscript{a} & 91.9 \\
    Interior designer  & Interior designers                                   & 82.8 \\
    Teacher            & Elementary and middle school teachers                & 79.2 \\
    Flight attendant   & Flight attendants                                    & 74.1 \\
    Newscaster         & News analysts, reporters, and journalists            & 41.1 \\
    Model              & Models, demonstrators, and product promoters         & ---\textsuperscript{b} \\
    Florist            & Floral designers                                     & ---\textsuperscript{b} \\
    Cheerleader        & (no BLS detailed occupation)                         & --- \\
    \midrule
    \multicolumn{2}{l}{\emph{All occupations (baseline)}}                     & 47.1 \\
    \bottomrule
  \end{tabular}

  \medskip
  \raggedright\footnotesize
  \textsuperscript{a}\,BLS row ``Secretaries and administrative assistants,
  except legal, medical, and executive.''
  \textsuperscript{b}\,BLS suppresses percent distributions for occupations
  with base employment below 50{,}000.
\end{table}

\subsection{Prompt-template ablation}
Fifteen of 67 occupations changed depicted gender between prompt styles;
thirteen flipped male~$\rightarrow$~female under descriptive phrasing and
only two flipped the other way (two-sided sign test, $p = 0.007$;
\Cref{tab:set3-prompt-ablation}). Descriptive role phrasing softens the
male default.

\begin{table}[t]
  \centering
  \small
  \caption{Prompt-style ablation in the initial set of generations without a specified gender. Each of the 67
  occupations was prompted twice. One prompt names the occupation
  (``portrait of a doctor''). One prompt describes the role without naming
  it (``someone who helps people in hospital''). Fifteen occupations
  changed depicted gender between prompt styles, and thirteen of the
  fifteen flips moved toward female under descriptive phrasing (two-sided
  sign test $p = 0.007$).}
  \label{tab:set3-prompt-ablation}
  \begin{tabular}{l r}
    \toprule
    \textbf{Quantity} & \textbf{Value} \\
    \midrule
    Occupations tested                                & 67 \\
    Same depicted gender under both prompt styles     & 52 \\
    Flipped male $\rightarrow$ female (descriptive)   & 13 \\
    Flipped female $\rightarrow$ male (descriptive)   & 2 \\
    Sign test (two-sided)                             & $p = 0.007$ \\
    \bottomrule
  \end{tabular}

  \medskip
  \raggedright\footnotesize
  Flips toward female under descriptive prompts were actor, architect,
  astronaut, author, baker, biker, biologist, firefighter, ice-cream
  vendor, journalist, painter, singer and student. Flips toward male were
  florist and interior designer.
\end{table}

\begin{figure}[t]
  \centering
  \includegraphics[height=0.82\textheight,keepaspectratio]{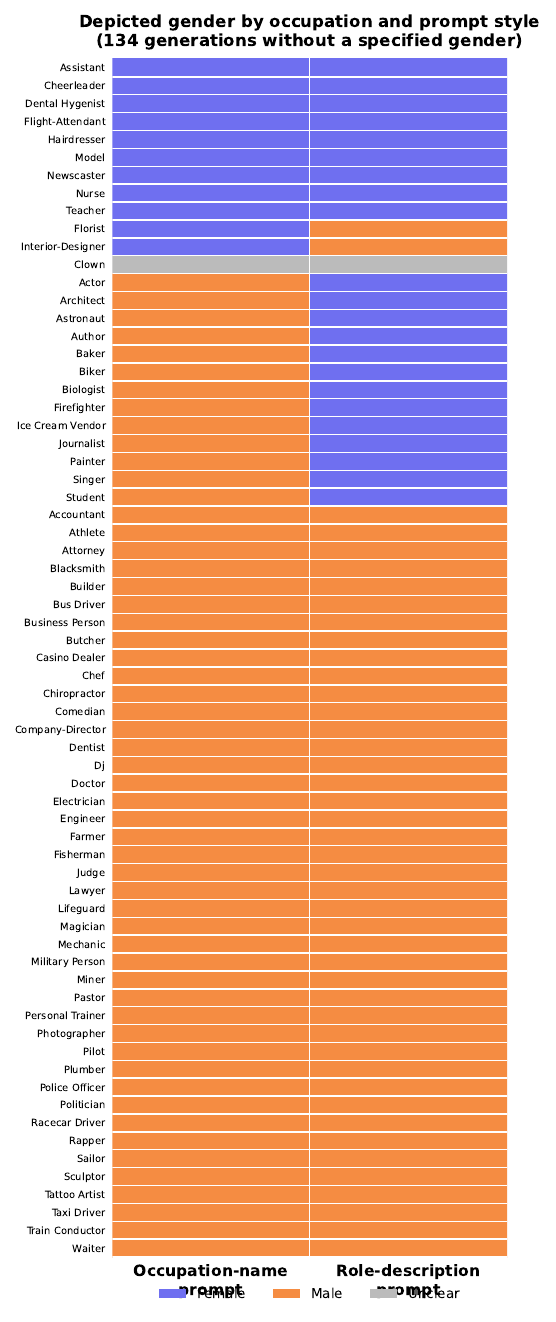}
  \caption{Midjourney defaults to male across occupations, and
  descriptive prompts soften the default. Depicted gender for each of 67
  occupations in the initial set of generations without a specified gender, under two gender-neutral prompt styles. One style names the occupation
  and the other describes the role without naming it. Rows are sorted by outcome. Occupations rendered female under
  occupation-name prompting cluster among female-majority occupations in
  U.S.\ labor statistics (\Cref{tab:set3-stereotype-index}).}
  \label{fig:set3-heatmap}
\end{figure}

\section{Partially AI-Generated Images}\label{sec:partial}

The AI-edited subset comprises 27 partially manipulated images. Each was
made from a real photograph in the dataset of \citet{kamali2025characterizing} with Midjourney inpainting, which
removes, modifies or adds content and leaves the rest of the scene
intact. The results are released as derivative works. The pairwise analyses below cover the 25 pairs with aligned source photographs, and edit-region statistics appear in \Cref{tab:set5-edit-area}.

We coded each of the 25 pairs manually into one of five mutually exclusive edit types by comparing the source photograph with the edited image and its pixel-difference map (\Cref{tab:set5-edit-type}). Replacing one person with another person is the most common edit (9 of 25), followed by background replacement (6) and object insertion or replacement (5), and four edits replace a person with an object or an animal. One edit re-renders a hat and a pair of sunglasses without removing the sunglasses that its prompt asked to remove. Edits that replace a person with an object or an animal and edits that replace the background alter the largest share of pixels, and person-to-person replacements alter a median of 12.7\% of pixels because the surrounding scene is kept.

\begin{table}[t]
  \centering
  \small
  \caption{AI-edited subset edit-region statistics over the 25 characterized real--edited pairs (the full release contains 27 edited images).
  Edited-pixel percentage uses a per-pixel absolute-difference threshold of
  25/255 after aligning resolutions; SSIM is computed globally.}
  \label{tab:set5-edit-area}
  \begin{tabular}{l r r r}
    \toprule
    \textbf{Measure} & \textbf{Median} & \textbf{Min} & \textbf{Max} \\
    \midrule
    Edited pixels (\%) & 14.3 & 1.3 & 53.9 \\
    Edit bounding box (\% of image) & 60.4 & 17.9 & 100.0 \\
    SSIM (source vs.\ edit) & 0.788 & 0.469 & 0.988 \\
    \bottomrule
  \end{tabular}
\end{table}

\begin{table}[t]
  \centering
  \small
  \caption{AI-edited subset edit-type taxonomy over the 25 characterized real--edited pairs. Each pair was coded manually into one category by inspecting the source photograph, the edited image and their pixel-difference map. Replacing one person with another person is the most common edit, and the median share of altered pixels varies by category.}
  \label{tab:set5-edit-type}
  \begin{tabular}{l r r}
    \toprule
    \textbf{Edit type} & \textbf{Pairs} & \textbf{Median edited pixels (\%)} \\
    \midrule
    Person replaced with another person      & 9 & 12.7 \\
    Background replaced                      & 6 & 23.2 \\
    Object inserted or replaced              & 5 & 9.8 \\
    Person replaced with an object or animal & 4 & 26.3 \\
    Local attribute re-rendered              & 1 & 1.3 \\
    \midrule
    \textbf{Total}                           & \textbf{25} & 14.3 \\
    \bottomrule
  \end{tabular}
\end{table}

\subsection{Edit-region size vs.\ detection difficulty}
\begin{figure}[t]
  \centering
  \includegraphics[width=0.65\linewidth]{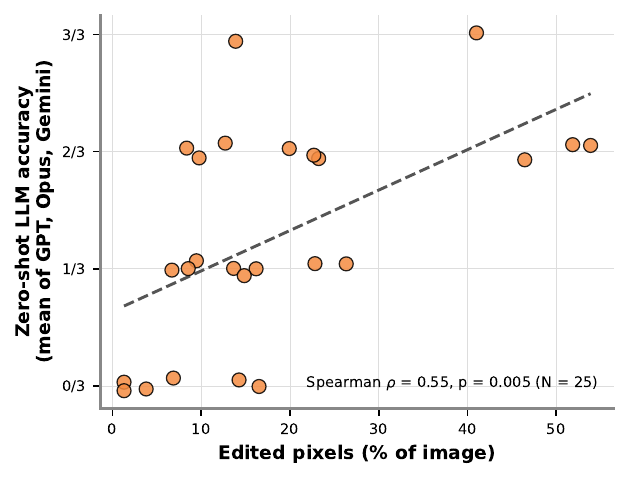}
  \caption{Larger edit regions are associated with higher zero-shot
  classification accuracy. Each point is one partially manipulated image in the AI-edited subset
  ($N = 25$); the x-axis gives the percent of pixels altered relative to the
  source photograph and the y-axis the mean zero-shot classification accuracy
  across three LLMs (GPT-4.1, Claude Opus, Gemini Pro). Edit-region size is
  positively associated with accuracy (Spearman $\rho = 0.55$, permutation
  $p = 0.005$).}
  \label{fig:set5-edit-vs-difficulty}
\end{figure}

We find that edit-region size is positively associated with zero-shot classification accuracy
(Spearman $\rho = 0.55$, permutation $p = 0.005$, $N = 25$;
\Cref{fig:set5-edit-vs-difficulty}). Edits altering a median of 14.3\% of pixels span the range of zero-shot
accuracy, from 0 to 3 of 3 models correct. The analysis covers 25 images
and three language models.

\section{Everyday Social Videos}\label{sec:video}

The Video subset comprises 56 AI-generated short-form videos across six
subject categories (\Cref{tab:set6-subjects}). The videos were produced
with Google Veo, OpenAI Sora and Kuaishou Kling. The generator split
appears in \Cref{tab:composition}. Subject-category coverage appears in \Cref{tab:set6-subjects}, and per-video technical characteristics appear in \Cref{tab:set6-metadata}. The per-video measurements below cover a 20-stimulus characterization subset.

\begin{table}[t]
  \centering
  \small
  \caption{Video-subset subject-category coverage for the 20-stimulus characterization subset (the full release contains 56 AI videos; the generator split appears in \Cref{tab:composition}).}
  \label{tab:set6-subjects}
  \begin{tabular}{l r}
    \toprule
    \textbf{Subject category} & \textbf{AI-generated videos} \\
    \midrule
    Physical activity       & 3 \\
    Street interviews       & 3 \\
    Protest and documentary & 3 \\
    Public speaking         & 3 \\
    Social media            & 5 \\
    Virtual communication   & 3 \\
    \midrule
    \textbf{Total}          & \textbf{20} \\
    \bottomrule
  \end{tabular}
\end{table}

\begin{table}[t]
  \centering
  \small
  \caption{Video-subset per-video technical characteristics for the characterization subset (measured with
  \texttt{ffprobe}). Median durations are 8.0\,s for both AI-generated videos and
  real references. AI-generated videos span a wider range of
  aspect ratios; real clips are either $960 \times 540$ or $360 \times 360$.}
  \label{tab:set6-metadata}
  \begin{tabular}{l r r}
    \toprule
    \textbf{Measure} & \textbf{Sora (AI, $n{=}20$)} & \textbf{Real ($n{=}20$)} \\
    \midrule
    Duration, median [range] (s) & 8.0 [5.7--10.0] & 8.0 [6.6--8.5] \\
    Frame rate (fps)             & 24, 30          & 24--60 \\
    Resolutions (distinct)       & 10 (528--1280\,px wide) & 2 ($960{\times}540$, $360{\times}360$) \\
    Audio track present          & 20/20           & 20/20 \\
    \bottomrule
  \end{tabular}
\end{table}

\subsection{Sora prompt complexity}

\begin{lstlisting}[caption={Representative Sora prompt for Video-subset generations.}]
youtube video style POV: a man teaching how to make pasta in
a home kitchen, handheld camera, natural lighting
\end{lstlisting}

\section{Discussion}\label{sec:discussion}

HumanCentric-Scenes-AI releases 296 AI-generated stimuli across four
domains of human-centric media. In our analyses, Midjourney CCTV-style images are sharper than the real
frames they mimic. Depicted gender under gender-neutral prompts is
associated with the gender composition of occupations. Zero-shot
classification accuracy is lower for smaller edits. The released metadata lets other researchers test these regularities on the full subsets.

\paragraph{Limitations.}
Several statistics are computed on 25-stimulus characterization subsets,
and the same scripts can be applied to the full subsets. Every difficulty
estimate also ages as generators improve, and the release records the
generator of each stimulus so future work can measure that decay.

\paragraph{Ethics and licensing.}
The AI-generated assets, the metadata, and the curation are released under
CC BY-NC 4.0. The partially
manipulated images in the AI-edited subset contain AI-edited regions over
public-source photographs; we release these as derivative works under a
fair-use rationale for academic research. People depicted in the Surveillance, Occupations and Video subsets are
synthetic, and the AI-edited subset retains the unaltered regions of its
public-source photographs. The dataset must \emph{not} be used
to train generative models.

\bibliographystyle{plainnat}
\bibliography{references}

\appendix

\section{Datasheet}\label{app:datasheet}
Following the datasheet framework of Gebru et al.~\citep{gebru2021datasheets}.

\paragraph{Motivation.} The dataset was created to support human-perception research on
AI-generated media. It was curated by the authors.

\paragraph{Composition.} 296 AI-generated stimuli in four named subsets (\Cref{tab:composition}). Every stimulus carries the metadata of
\Cref{tab:metadata-schema}. People
in the Surveillance, Occupations and Video subsets are synthetic. The
AI-edited subset is derived from public-source photographs of real scenes
with an altered region, and no personally identifying information is
intentionally added.

\paragraph{Collection process.} Surveillance and Occupations images were
generated with Midjourney from the prompts released per stimulus; Occupations
follows a balanced paired design covering 64 occupations with one female
and one male version of each. Video stimuli were generated with Veo, Sora and Kling from
category-matched prompts. AI-edited
stimuli apply AI edits to regions of public-source photographs and are released
as derivative works.

\paragraph{Preprocessing.} Files are renamed to content hashes;
curation-time descriptive names are preserved in \texttt{original\_filename}.
Surveillance-subset stimuli in the characterization subset were normalized to
$600 \times 600$ pixels.

\paragraph{Uses.} The intended use is human-perception research on
AI-generated media. Calibration of AI-detection classifiers, bias analysis of
generative models and prompt-template ablation are also supported. Prohibited
uses are training generative models and any use implying the depicted
synthetic persons are real.

\paragraph{Distribution and license.} Hosted on Hugging Face
(\url{https://huggingface.co/datasets/Nima0Kamali/humancentric-scenes-ai})
under CC BY-NC 4.0 for the AI-generated assets, metadata, and curation.

\paragraph{Maintenance.} Maintained by the authors; the corresponding author
is the point of contact. The Midjourney version is recorded in the prompt for 32 of the 85 Surveillance images and is not recorded for the other stimuli. The Video subset's Veo stimuli are pending restoration to the repository's main branch, and their generator labels in \texttt{metadata.csv} read gemini and need correction to veo.

\section{Example Stimuli}\label{app:examples}

\begin{figure}[h]
  \centering
  \includegraphics[width=0.88\linewidth,height=0.82\textheight,keepaspectratio]{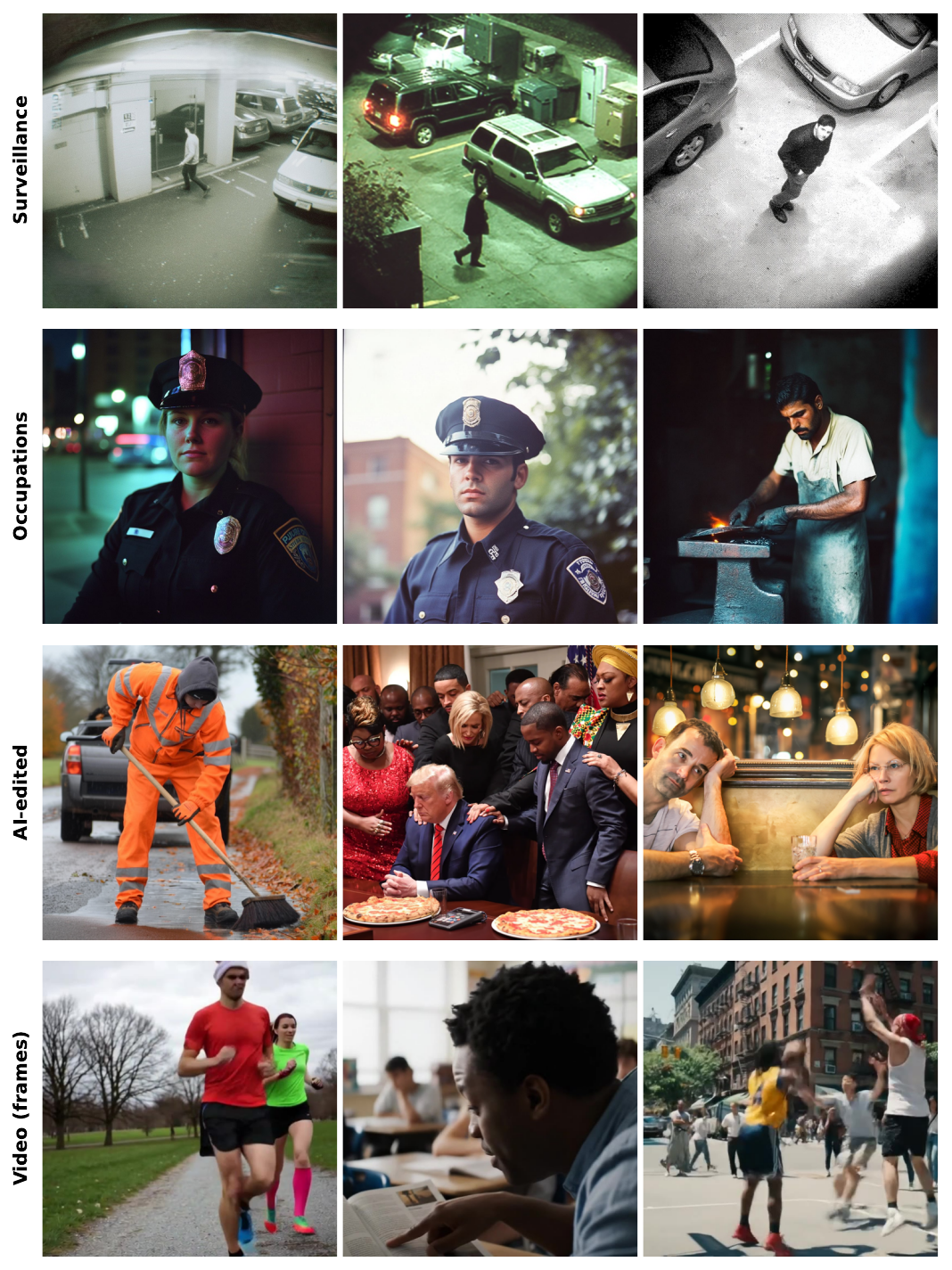}
  \caption{Example AI-generated stimuli from each subset (top to bottom:
  Surveillance, Occupations, AI-edited, Video frames). All stimuli shown are
  AI-generated.}
  \label{fig:example-grid}
\end{figure}

\end{document}